\documentclass[]{spie}  

\usepackage{amsmath,amsfonts,amssymb}
\usepackage{graphicx}
\usepackage[colorlinks=true, allcolors=blue]{hyperref}
\usepackage{array}

\title{PAHD: Perception-Action based Human Decision Making using Explainable Graph Neural Networks on SAR Images}

\author[a]{Sasindu Wijeratne}
\author[a]{Bingyi Zhang}
\author[b]{Rajgopal Kannan}
\author[a]{Viktor Prasanna}
\author[b]{Carl Busart}
\affil[a]{University of Southern California, Los Angeles, CA}
\affil[b]{DEVCOM US Army Research Lab, Playa Vista, CA}

\authorinfo{Further author information: (Send correspondence to Sasindu Wijeratne)\\
Sasindu Wijeratne: kangaram@usc.edu\\ 
Bingyi Zhang: bingyizh@usc.edu\\
Rajgopal Kannan: rajgopal.kannan.civ@army.mil\\
Viktor Prasanna: prasanna@usc.edu\\
Carl Busart: carl.e.busart.civ@army.mil
}

\begin{document} 
\maketitle

\begin{abstract}
Synthetic Aperture Radar (SAR) images are commonly utilized in military applications for automatic target recognition (ATR). Machine learning (ML) methods, such as Convolutional Neural Networks (CNN) and Graph Neural Networks (GNN), are frequently used to identify ground-based objects, including battle tanks, personnel carriers, and missile launchers. Determining the vehicle class, such as the BRDM2 tank, BMP2 tank, BTR60 tank, and BTR70 tank, is crucial, as it can help determine whether the target object is an ally or an enemy.
While the ML algorithm provides feedback on the recognized target, the final decision is left to the commanding officers. Therefore, providing detailed information alongside the identified target can significantly impact their actions. This detailed information includes the SAR image features that contributed to the classification, the classification confidence, and the probability of the identified object being classified as a different object type or class.
We propose a GNN-based ATR framework that provides the final classified class and outputs the detailed information mentioned above. This is the first study to provide a detailed analysis of the classification class, making final decisions more straightforward. Moreover, our GNN framework achieves an overall accuracy of 99.2\% when evaluated on the MSTAR dataset, improving over previous state-of-the-art GNN methods.
\end{abstract}


\section{INTRODUCTION}
\label{sec:intro}
The increased use of machine learning (ML) has created a growing demand for explainability~\cite{8466590}, particularly in high-stakes decision-making fields like synthetic aperture radar (SAR)~\cite{https://doi.org/10.48550/arxiv.2204.06783, Huang_2022} image-based automatic target recognition (ATR). In the past, domain experts designed SAR image analysis systems that comprised statistical classifiers using handcrafted image properties~\cite{rs11161942} like edges and corners to perform specific tasks. However, ML learns these features from data to optimize results. In SAR image-based ATR, deep learning techniques like convolutional neural networks (CNNs)~\cite{DBLP:journals/corr/SzegedyLJSRAEVR14} and graph neural networks (GNNs)~\cite{10.5555/3294771.3294869} have been successful in identifying objects on the ground, including battle tanks, personnel carriers, and missile launchers. Identifying the vehicle class is also crucial, as it helps to determine whether the target object is an ally or an enemy.

While ML algorithms provide feedback on the recognized target, the final decision is left to commanding officers~\cite{article1012}. Therefore, detailed information alongside the identified target can significantly impact their actions. This detailed information includes the SAR image features that contributed to the classification, the classification confidence, and the probability of the identified object being classified as a different object type or class.

GNNs are complex, consisting of multiple layers connected via many nonlinear intertwined relations~\cite{https://doi.org/10.48550/arxiv.2109.10119}. It is challenging to fully comprehend how the GNN decides, making it a black box~\cite{rai2020explainable}. The concern is mounting in various fields of application that these black boxes may be biased and that such bias goes unnoticed, with far-reaching consequences in military applications. Therefore, a call has been made for explainable artificial intelligence (Explainable AI) to better understand the black box.

In this work, we propose a GNN-based ATR framework that provides not only the final classified class but also outputs details on the confidence in the classifications and position of interests. Our proposed framework represents the first study to provide a detailed analysis of the classification class, simplifying the decision-making process for commanding officers. We focus on GNNs as they can effortlessly satisfy SWaP constraints to support edge-military applications.

To address the demand for explainability, we propose a model-specific explainable AI model~\cite{8400040} that is limited to our GNN model. We use attributes specific to our proposed GNN model, such as input graphs and layer shapes. Our proposed GNN-based ATR framework provides an efficient and accurate approach for target recognition in SAR images. With the inclusion of explainability, it helps to ensure fairness and transparency, particularly in high-stakes decision-making domains like military applications. It also achieves an overall accuracy of 99.2\% in the MSTAR dataset, improving over previous state-of-the-art GNN methods.


\section{EXPLAINABLE GNN MODEL}

   \begin{figure} [ht]
   \begin{center}
   \begin{tabular}{c} 
   \includegraphics[width=0.80\linewidth]{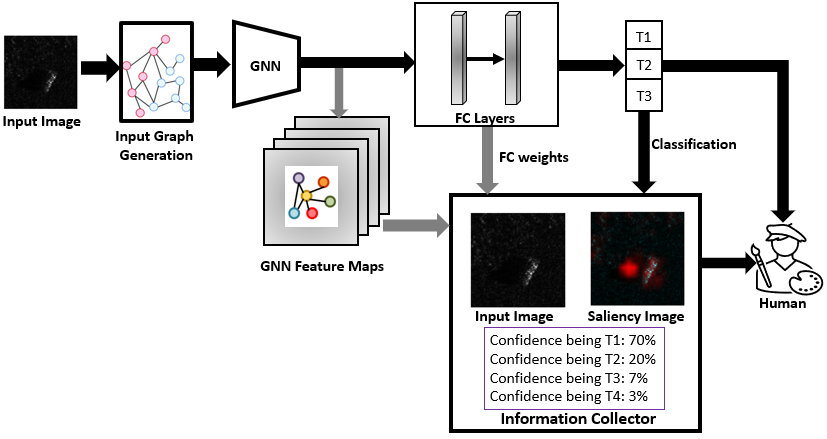}
   \end{tabular}
   \end{center}
   \caption[example] 
   { \label{fig:explain_GNN} 
Explainable GNN Framework for SAR ATR}
   \end{figure} 

The primary focus of our work is to develop an Explainable Graph Neural Network (GNN) framework for Automatic Target Recognition (ATR) in Synthetic Aperture Radar (SAR) images. This framework, which comprises several stages, is illustrated in Figure~\ref{fig:explain_GNN}.
\\\\
\textbf{Input Graph Generation:}
We represent a SAR image as a graph, denoted as $\mathcal{G}(\mathcal{V},\mathcal{E}, X^{0})$, where each pixel is viewed as a vertex in the graph. To maintain the structural information of the target image, we connect each vertex/pixel to its 8 neighbors (horizontal, vertical, and diagonals) with unweighted edges. The input features of a vertex set, i.e., $X^{0}$, include the pixel's grayscale value. This feature set is forwarded to the GNN as separate input channels. Since each pixel in the image is mapped to a vertex, the generated graph has the exact dimensions as the original image.
\\\\
\begin{table}[ht]
\caption{Notations used in GNNs}
\begin{center}
\vspace{-3mm}
\resizebox{\columnwidth}{!}{%
\begingroup
\setlength{\tabcolsep}{6pt} 
\renewcommand{\arraystretch}{1.5} 
\begin{tabular}{|l c|l c|}
 \hline
 \textbf{Notation} & \textbf{Description} & \textbf{Notation} & \textbf{Description} \\
 \hline\hline
 $\mathcal{G}(\mathcal{V},\mathcal{E},X^{0})$ & input graph & $\mathcal{V}$ & set of vertices \\ 
 \hline
 $\mathcal{E}$ & set of edges & $v_i$ & $i^{th}$ vertex \\
\hline
 $e_{i,j}$ & edge from $v_i$ to $v_j$ & $L$ & number of GNN layers \\
\hline
$h^{l}_{i}$ & feature set of $v_i$ at layer $l$ & $\mathcal{N}(i)$ & neighbours of $v_i$ \\
 \hline
\end{tabular}%
\endgroup
}
\label{notations}
\end{center}
\end{table}
\textbf{GNN:}
To learn from the structural information of the input graph, we utilize Graph Neural Networks (GNNs), which can embed information in a low-dimensional vector representation or graph embedding. GNNs follow the message-passing paradigm, where vertices recursively aggregate information from their neighbors. This work uses the GraphSAGE~\cite{10.5555/3294771.3294869} GNN model for a graph classification task. For an input graph $\mathcal{G}(\mathcal{V},\mathcal{E}, X^{0})$, the GraphSAGE follows the aggregate-update paradigm, as shown in Equation~\ref{eq:GNN}. The notations are defined in Table~\ref{notations}.

\begin{equation} \label{eq:GNN}
\begin{aligned}
  &\text{Aggregate: } z^{l}_{i} = Mean(h^{l-1}_{j}: j \in \mathcal{N}(i) \cup {i}) \\
  &\text{Update: } h^{l}_{i} = RELU((z^{l}_{i}W^{l}_{\text{neighbour}} + b^{l}_{\text{neighbour}})(h^{l-1}_{i}W^{l}_{\text{self}} + b^{l}_{\text{self}}))
\end{aligned}
\end{equation}

Each GNN layer includes a GraphSAGE layer~\cite{10.5555/3294771.3294869} following RELU, Max Pool, and attention layer.

Since the input graph has a 2-D grid structure, we adopt a similar pooling strategy as the Convolutional Neural Network (CNN) for the 2-D image. Within each local $s\times s$ range, which has $s^2$ vertices, we perform the max pooling operator on the $s^2$ vertices to obtain an output vertex. The attention module used in each layer consists of a Channel Attention module and a Spatial Attention module similar to the CNN attention mechanism~\cite{https://doi.org/10.48550/arxiv.1807.06521}. We perform weight pruning by training the model using lasso regression~\cite{tibshirani1996regression} to reduce the total computation complexity.

\hspace{-5mm}\textbf{GNN Feature Maps:}
GNN feature maps identify the output graphs' vertices that mainly contribute to the classification task in each GNN layer. After identifying these vertices, they are extrapolated to the input graph and remapped to the pixel level by executing the reverse process to generate input graphs from the image.
\\\\
\textbf{FC Layers:}
Fully connected layers are used to take every output vertices of the last layer of GNN and predict the probability of the input image belonging to a specific class using a multi-layer perceptron (MLP).
\\\\
\textbf{Information Collector:}
The information collector collects and shows (1) GNN feature maps and visualizes the most critical pixels for the classification process and (2) the top N probabilities of the input SAR image belonging to the classification classes to the decision maker.

\begin{table}[h!]
\caption{Sample images and their classification confidence}\label{tbl:explainAI}
  \centering
  \begin{tabular}{| c | c |}
    \hline
    Explainable Figure & Confidence \\ \hline
    \begin{minipage}{.3\textwidth}
      \includegraphics[width=\linewidth, height=58mm]{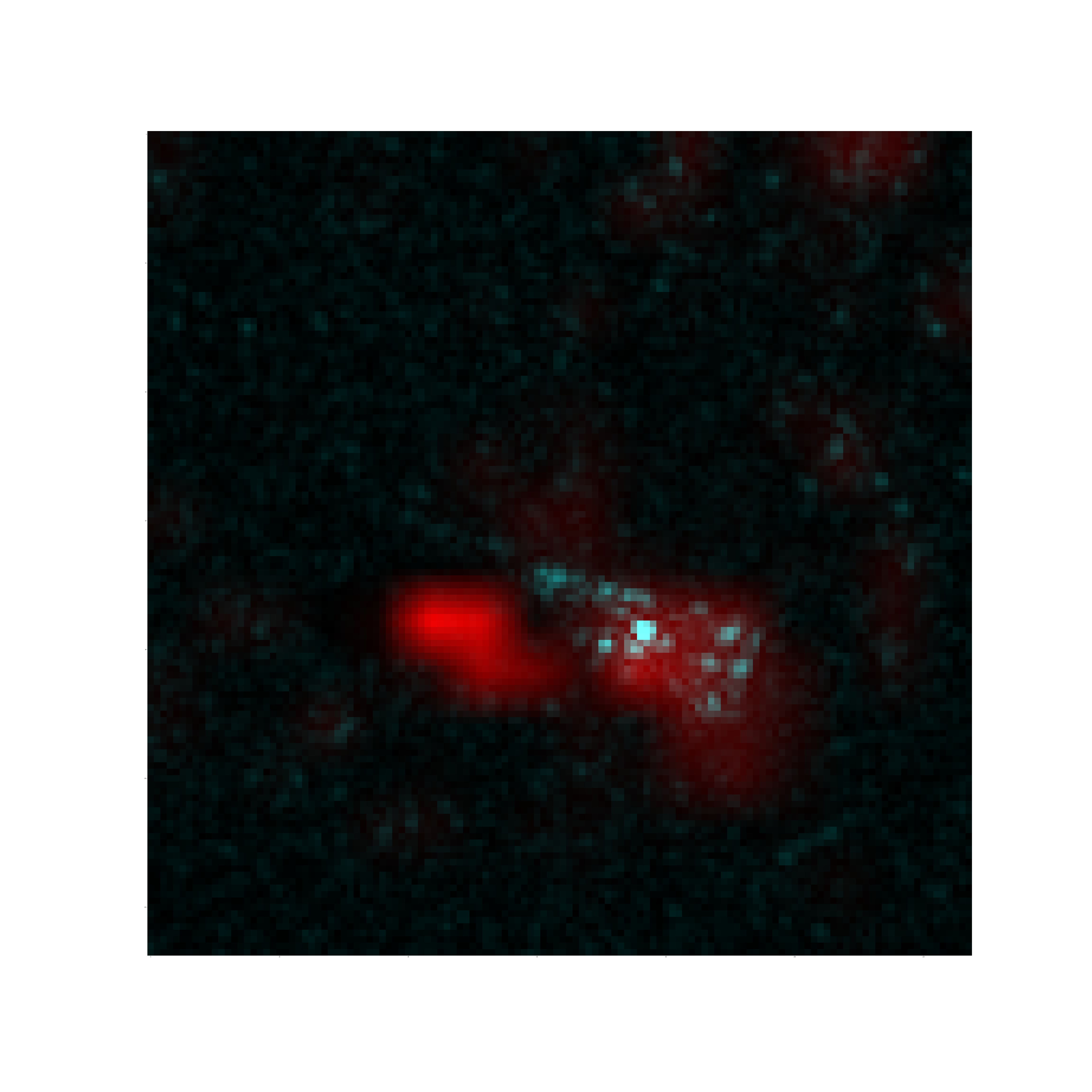}
            \center Original Class: 2S1
    \end{minipage}
    &
    \begin{minipage}{5cm}
      \begin{itemize}
        \item 2S1: 52.61\%
        \item ZSU234: 33.52\%
        \item T62: 6.33\%
        \item BTR70: 4.47\%
      \end{itemize}
    \end{minipage}
    \\ \hline
    \begin{minipage}{.3\textwidth}
      \includegraphics[width=\linewidth, height=58mm]{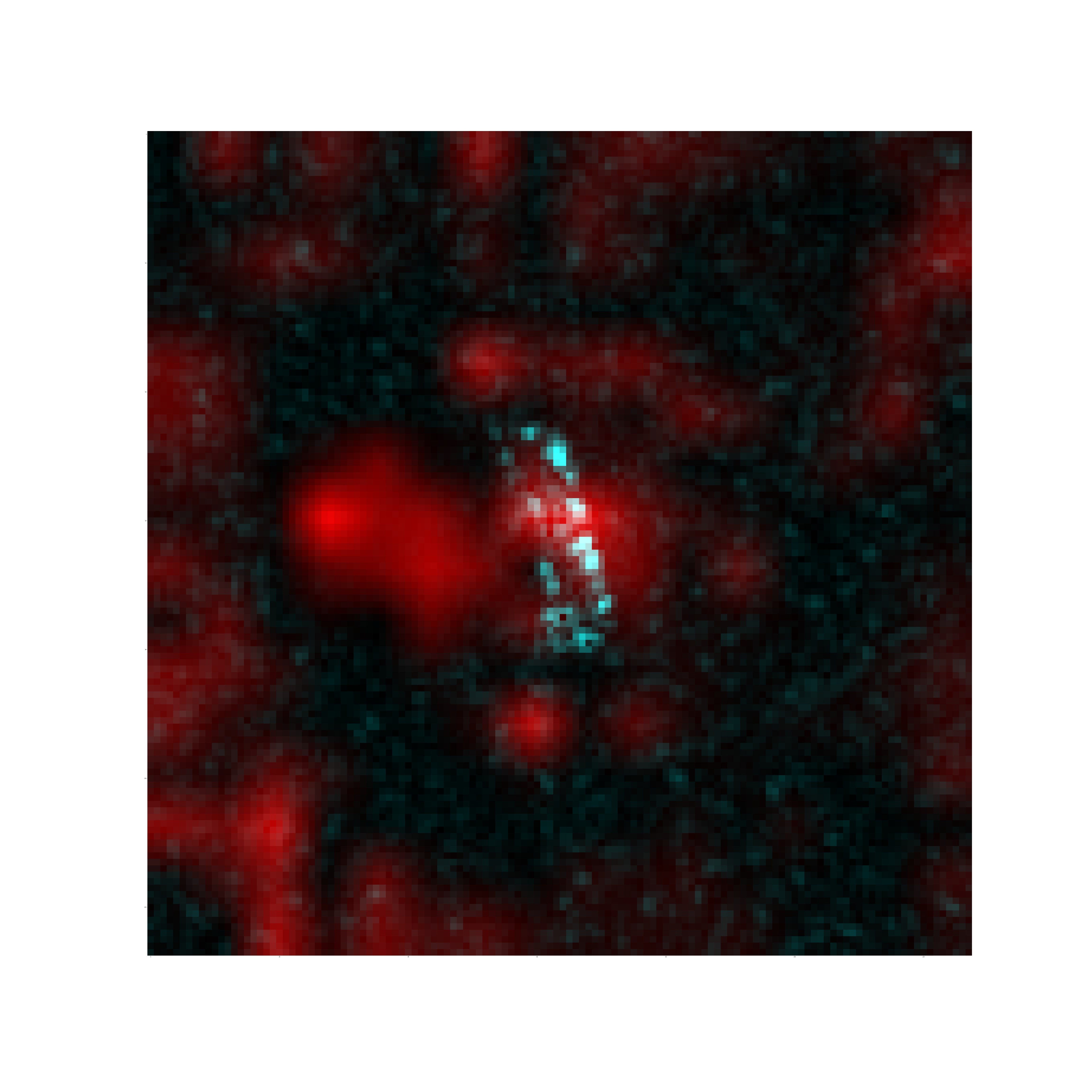}
      \center Original Class: BTR70
    \end{minipage}
    &
    \begin{minipage}{5cm}
      \begin{itemize}
        \item BMP2: 46.41\%
        \item BTR70: 39.23\%
        \item BTR-60: 6.82\%
        \item T72: 4.23\%
      \end{itemize}
    \end{minipage}
    \\ \hline
    \begin{minipage}{.3\textwidth}
      \includegraphics[width=\linewidth, height=58mm]{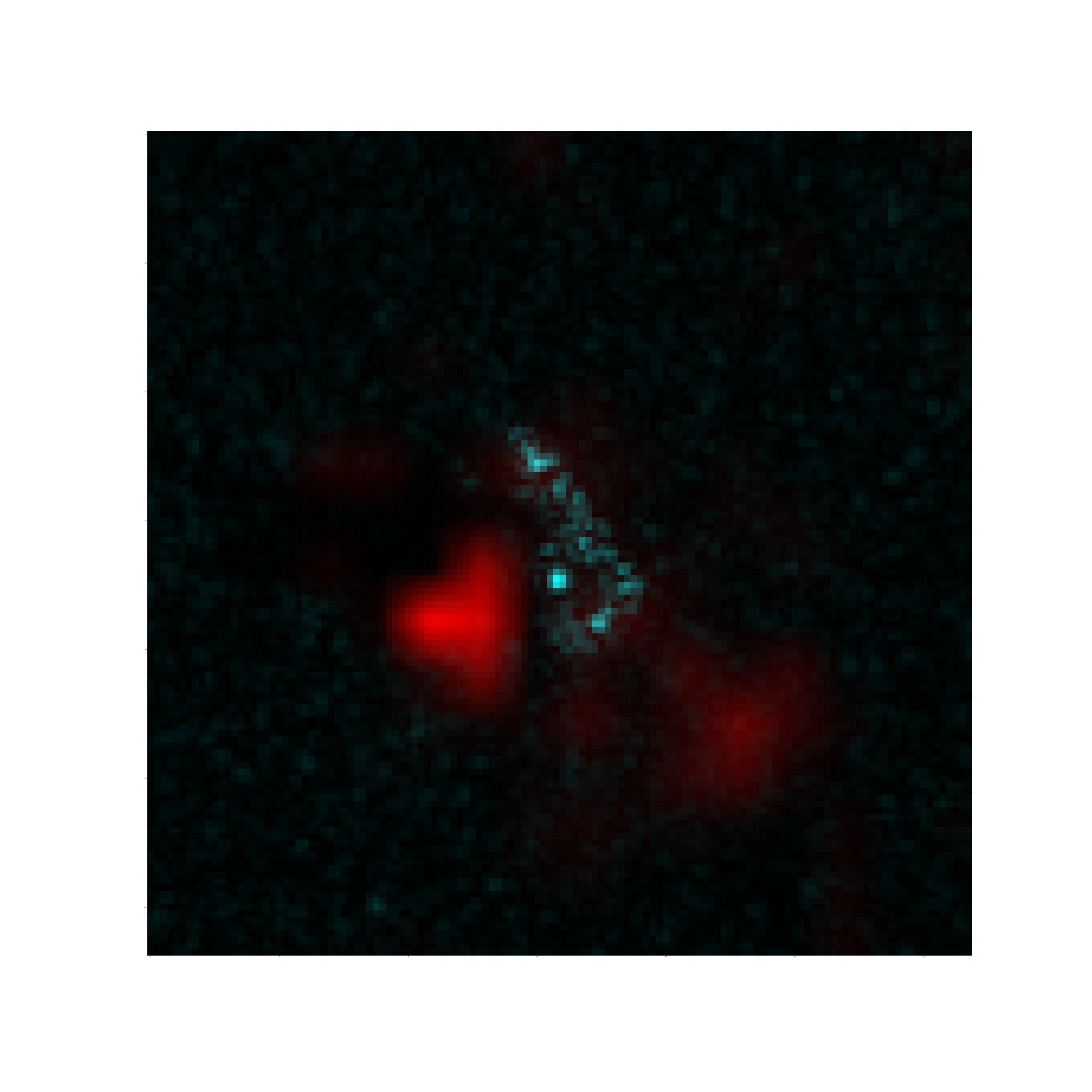}
      \center Original Class: T72
    \end{minipage}
    &
    \begin{minipage}{5cm}
      \begin{itemize}
        \item BRDM2: 26.64\%
        \item T72: 25.42\%
        \item BTR70: 16.99\%
        \item BTR-60: 14.58\%
      \end{itemize}
    \end{minipage}
    \\ \hline
  \end{tabular}
\end{table}

\section{EVALUATION}

We conduct experiments using the widely used MSTAR dataset~\cite{MSTAR0}. The dataset contains the SAR images of 10 classes of ground vehicles. Each image has a size of $128\times 128$, and each pixel has a grayscale value. The setup used in this work follows the same experiment setup used in state-of-the-art work in the SAR ATR domain.


\subsection{Baselines \& Comparisons}\label{baselines}
We use 5 baseline ML models to compare our work. Gabor + SVM \cite{5367456} introduces SAR target recognition based on Gabor filter bank sub-block statistical feature extraction. This work uses pre-determined configurations for Gabor Filters, while our work uses adaptive Gabor Filters, which can tune the parameters. A lightweight attention mechanism combined with a CNN model is proposed in CNN + Attention~\cite{9246598} for accurately classifying SAR images. We use a similar attention mechanism proposed in this work in  GNN layers to increase the accuracy of the GNN model used in our work. CNN~\cite{10.1117/12.2176558} explores the performance of a Deep CNN approach to the classification of SAR imagery using the MSTAR public release dataset. TAI-SARNET~\cite{2020Senso..20.1724Y} propose an effective lightweight CNN model incorporating transfer learning to handle SAR targets recognition tasks while dealing with speckle noise embedded in original SAR images. Multi-view~\cite{8207785} proposes a new approach for SAR ATR, in which a multiview deep learning framework where Based on the multiview SAR ATR pattern, which guarantees a large number of inputs for network training without needing many raw SAR images while using a CNN containing a parallel network topology with multiple inputs.

Table~\ref{tab:speed-up} shows the comparison of accuracy between our work and the baselines discussed in Section~\ref{baselines}. According to the results, our work achieves the highest accuracy (i.e., 99.2\%) under the same experiment setup.

\begin{table}
  \vspace{-4mm}
  \caption{Comparison of accuracy of different ML models}
    \vspace{4mm}
  \centering
  \begin{tabular}{|c|c|c|c|c|c|}
    \hline
Gabor + SVM~\cite{5367456} & CNN + Attention~\cite{9246598} & CNN~\cite{10.1117/12.2176558} & TAI-SARNET~\cite{2020Senso..20.1724Y} & Multi-view~\cite{8207785} &  Our work \\
    \hline
 93.50\% & 99.30\% & 92.35\% & 97.97\% & 98.52\% & \textbf{99.20\%} \\
    \hline
  \end{tabular}
  \label{tab:speed-up}
\end{table}

\begin{table}
  \caption{Comparison of accuracy of different ML models}
  \centering
  \begin{tabular}{|c|c|c|c|c|}
    \hline
SVM [2009] & CNN [2020] & Multi-view & CNN + Attention [2022] &  GNN \\
    \hline
 93.50\% & 97.97\% & 98.52\% & 92.35\% & \textbf{99.20\%} \\
    \hline
  \end{tabular}
  \label{tab:speed-up}
\vspace{-3mm}
\end{table}



\subsection{Explainable Results}

   \begin{figure} [ht]
   \begin{center}
   \begin{tabular}{c}
   \includegraphics[width=0.50\linewidth]{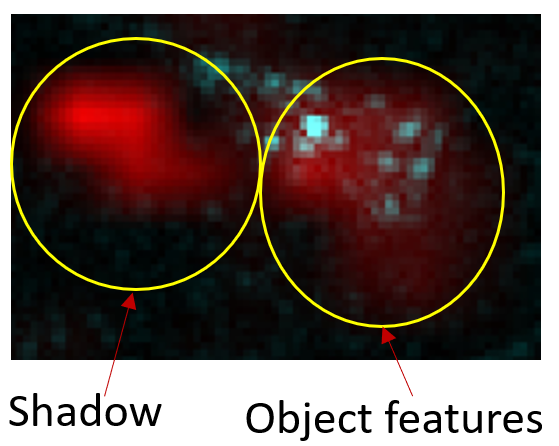}
   \end{tabular}
   \end{center}
   \caption[example] 
   { \label{fig:exracted} 
Main regions in the image contributed to the classification}
   \end{figure} 

The framework consists of an Information Collector module that presents the results of the classifier to the decision-makers. The classifier's accurate classifications are primarily based on the object's features and the shape of the shadow generated by reflected radar signals, as depicted in Figure~\ref{fig:exracted}. The highly weighted pixels for classification are shown in red. However, we have observed that in cases of misclassification, the classifier relies on background information. Therefore, we suggest that decision-makers should not depend solely on the classifier's outcome in such instances and consider the potential influence of background information in their decision-making process.
Table~\ref{tbl:explainAI} shows example classification results extracted from the Information Collector. Similar to Figure~\ref{fig:exracted}, highly weighted pixels for classification are shown in red. In Figure 1, a 2S1 is correctly classified with higher confidence, where the tank's pixels and its shadow are used for classification. In Figure 2, the BTR70 is misclassified as BMP2. The classifier's highly weighted pixels are scattered around the background, indicating that background noise significantly influences the final classification results. Despite the misclassification, the confidence level of classification is high, showing that background noise greatly impacted the classifier. In Figure 3, the T72 is misclassified as BRDM2. The background noise contributes to the classification, as there are insignificant red pixels on the T72 object, suggesting that the object's features have a negligible contribution to the classification.

\section{CONCLUSION}

SAR images are widely used in military applications for automatic target recognition using ML methods. The classification of ground-based objects, such as tanks and missile launchers, is crucial for determining whether the target object is an ally or an enemy. However, the final decision is left to the commanding officers, and detailed information alongside the recognized target can significantly impact their final action.
In this study, we proposed a GNN-based ATR framework that provides the final classified class and outputs detailed information on SAR image features that contributed to the classification. Our proposed framework achieves an overall accuracy of 99.20\% in the MSTAR dataset, with a smaller model size and less computation cost than state-of-the-art CNNs.

Furthermore, our study revealed that correct classifications are mainly based on the features of the object and its shadow, while misclassifications often rely on background information. Therefore, we suggest that decision-makers avoid depending solely on the outcome of misclassifications but take into consideration the potential influence of background information in their decision-making process.

Overall, our proposed GNN-based ATR framework with detailed information on classification results can provide decision-makers with more accurate and reliable information to support their decision-making processes.

\section{ACKNOWLEDGEMENT}
This work is supported by DEVCOM Army Research Lab (ARL) grant W911NF2220159.

\bibliography{report}

\begin{thebibliography}{10}

\bibitem{8466590}
Adadi, A. and Berrada, M., ``Peeking inside the black-box: A survey on
  explainable artificial intelligence (xai),'' {\em IEEE Access}~{\bf 6},
  52138--52160 (2018).

\bibitem{https://doi.org/10.48550/arxiv.2204.06783}
Su, S., Cui, Z., Guo, W., Zhang, Z., and Yu, W., ``Explainable analysis of deep
  learning methods for sar image classification,'' (2022).

\bibitem{Huang_2022}
Huang, Z., Yao, X., Liu, Y., Dumitru, C.~O., Datcu, M., and Han, J.,
  ``Physically explainable {CNN} for {SAR} image classification,'' {\em {ISPRS}
  Journal of Photogrammetry and Remote Sensing}~{\bf 190},  25--37 (aug 2022).

\bibitem{rs11161942}
Liu, X., He, C., Xiong, D., and Liao, M., ``Pattern statistics network for
  classification of high-resolution sar images,'' {\em Remote Sensing}~{\bf
  11}(16) (2019).

\bibitem{DBLP:journals/corr/SzegedyLJSRAEVR14}
Szegedy, C., Liu, W., Jia, Y., Sermanet, P., Reed, S.~E., Anguelov, D., Erhan,
  D., Vanhoucke, V., and Rabinovich, A., ``Going deeper with convolutions,''
  {\em CoRR}~{\bf abs/1409.4842} (2014).

\bibitem{10.5555/3294771.3294869}
Hamilton, W.~L., Ying, R., and Leskovec, J., ``Inductive representation
  learning on large graphs,'' in [{\em Proceedings of the 31st International
  Conference on Neural Information Processing
  Systems}{\nolinebreak\hspace{0.1em}]},  {\em NIPS'17},  1025–1035, Curran
  Associates Inc., Red Hook, NY, USA (2017).

\bibitem{article1012}
Kaempf, G., Klein, G., Thordsen, M., and Wolf, S., ``Decision making in complex
  naval command-and-control environments,'' {\em Human Factors}~{\bf 38},
  220--231 (06 1996).

\bibitem{https://doi.org/10.48550/arxiv.2109.10119}
Grassia, M., De~Domenico, M., and Mangioni, G., ``mgnn: Generalizing the graph
  neural networks to the multilayer case,'' (2021).

\bibitem{rai2020explainable}
Rai, A., ``Explainable ai: From black box to glass box,'' {\em Journal of the
  Academy of Marketing Science}~{\bf 48},  137--141 (2020).

\bibitem{8400040}
Došilović, F.~K., Brčić, M., and Hlupić, N., ``Explainable artificial
  intelligence: A survey,'' in [{\em 2018 41st International Convention on
  Information and Communication Technology, Electronics and Microelectronics
  (MIPRO)}{\nolinebreak\hspace{0.1em}]},   0210--0215 (2018).

\bibitem{https://doi.org/10.48550/arxiv.1807.06521}
Woo, S., Park, J., Lee, J.-Y., and Kweon, I.~S., ``Cbam: Convolutional block
  attention module,'' (2018).

\bibitem{tibshirani1996regression}
Tibshirani, R., ``Regression shrinkage and selection via the lasso,'' {\em
  Journal of the Royal Statistical Society: Series B (Methodological)}~{\bf
  58}(1),  267--288 (1996).

\bibitem{MSTAR0}
Tu, S., Su, Y., Wang, W., Xiong, B., and Li, Y., ``Automatic target recognition
  scheme for a high-resolution and large-scale synthetic aperture radar
  image,'' {\em Journal of Applied Remote Sensing}~{\bf 9},  096039 (06 2015).

\bibitem{5367456}
Hu, F.-m., Zhang, P., Yang, R.-l., and Fan, X.-h., ``Sar target recognition
  based on gabor filter and sub-block statistical feature,'' in [{\em 2009 IET
  International Radar Conference}{\nolinebreak\hspace{0.1em}]},   1--4 (2009).

\bibitem{9246598}
Zhang, M., An, J., Yu, D.~H., Yang, L.~D., Wu, L., and Lu, X.~Q.,
  ``Convolutional neural network with attention mechanism for sar automatic
  target recognition,'' {\em IEEE Geoscience and Remote Sensing Letters}~{\bf
  19},  1--5 (2022).

\bibitem{10.1117/12.2176558}
Morgan, D. A.~E., ``{Deep convolutional neural networks for ATR from SAR
  imagery},'' in [{\em Algorithms for Synthetic Aperture Radar Imagery
  XXII}{\nolinebreak\hspace{0.1em}]},  Zelnio, E. and Garber, F.~D., eds.,
  {\bf 9475},  94750F, International Society for Optics and Photonics, SPIE
  (2015).

\bibitem{2020Senso..20.1724Y}
{Ying}, Z., {Xuan}, C., {Zhai}, Y., {Sun}, B., {Li}, J., {Deng}, W., {Mai}, C.,
  {Wang}, F., {Labati}, R.~D., {Piuri}, V., and {Scotti}, F., ``{TAI-SARNET:
  Deep Transferred Atrous-Inception CNN for Small Samples SAR ATR},'' {\em
  Sensors}~{\bf 20},  1724 (Mar. 2020).

\bibitem{8207785}
Pei, J., Huang, Y., Huo, W., Zhang, Y., Yang, J., and Yeo, T.-S., ``Sar
  automatic target recognition based on multiview deep learning framework,''
  {\em IEEE Transactions on Geoscience and Remote Sensing}~{\bf 56}(4),
  2196--2210 (2018).

\end{thebibliography}
\bibliographystyle{spiebib}

\end{document}